\documentclass[english]{IEEEtran}
\usepackage[T1]{fontenc}
\usepackage[latin9]{inputenc}
\usepackage{color}
\usepackage{amsmath}
\usepackage{stackrel}
\usepackage{graphicx}

\makeatletter

\providecommand{\tabularnewline}{\\}

\makeatother

\usepackage{babel}
\begin{document}

\title{A Taxonomy for Neural Memory Networks}

\author{Ying Ma, Jose Principe, \textit{Life Fellow, IEEE} \thanks{The authors are with the Computational NeuroEngineering Laboratory,
University of Florida, Gainesville, FL 32611 USA (e-mail: mayingbit2011@gmail.com;
principe@cnel.ufl.edu).} }
\maketitle
\begin{abstract}
In this paper, a taxonomy for memory networks is proposed based on
their memory organization. The taxonomy includes all the popular memory
networks: vanilla recurrent neural network (RNN), long short term
memory (LSTM ), neural stack and neural Turing machine and their variants.
The taxonomy puts all these networks under a single umbrella and shows
their relative expressive power , i.e. vanilla RNN$\subseteq$ LSTM$\subseteq$neural
stack$\subseteq$neural RAM. The differences and commonality between
these networks are analyzed. These differences are also connected
to the requirements of different tasks which can give the user instructions
of how to choose or design an appropriate memory network for a specific
task. As a conceptual simplified class of problems, four tasks of
synthetic symbol sequences: counting, counting with interference,
reversing and repeat counting are developed and tested to verify our
arguments. And we use two natural language processing problems to
discuss how this taxonomy helps choosing the appropriate neural memory
networks for real world problem. 
\end{abstract}


\section{Introduction}

\label{sec:intro}

Memory has a pivotal role in human cognition and many different types
are well known and intensively studied\cite{fuster2015prefrontal}.
In neural networks and signal processing the use of memory is concentrated
in preserving in some form (by storing past samples or using a state
model) the information from the past. A system is said to include
memory if the system's output is a function of the current and past
samples. Feedforward neural networks are memoryless, but the time
delay neural network \cite{waibel1990phoneme}, the gamma neural model
\cite{de1992gamma} and recurrent neural networks are memory networks.
An important theoretical result showed that these networks are universal
in the space of myopic functions \cite{sandberg1997uniform}. A methodology
to quantify linear memories was presented in \cite{de1992gamma},
which proposed an analytic expression for the compromise between memory
depth (how much the past is remembered) and memory resolution (how
specifically the system remembers a past event). A similar compromise
exists for nonlinear dynamic memories (i.e. using nonlinear state
variables to represent the past), but is depends on the type of nonlinearity
and there is no known close form solution. It is fair to say that
currently the most utilized neural memory is the recurrent neural
networks (RNN) for sequence learning. Compared to the time delay neural
network, RNN keeps a processed version of the past signal in its state.
\cite{elman1990finding}\cite{jordan:attractor} proposed the first
classic version of RNN which introduces memory by adding a feedback
from the hidden layer output to its input for sequence recognition.
They are often referred to as vanilla RNN nowadays. A large body of
work also used RNNs for dynamic modeling of complex dynamical and
even chaotic systems \cite{haykin1998making}. However, although RNNs
are theoretically Turing complete if well-trained, they usually do
not perform well when the sequence is long. Long short term memory
(LSTM) \cite{hochreiter1997long} was proposed to provide more flexibility
to RNNs by\textcolor{red}{{} }\textcolor{black}{employing }an external
memory called cell state to deal with the vanish gradient problem.
\textcolor{black}{Three logic} gates are also introduced to adjust
this external memory and internal memory. Later on several variants
were proposed \cite{greff2017lstm}. One example needs to be mentioned
is the Gated Recurrent Unit \cite{cho2014learning}which combines
the forget and input gates and makes some other changes to make the
model simpler. Another variant is the peephole network \cite{gers2000recurrent}
which makes the gate layers look not only the input and hidden state
but also the cell state. With the help of the external memory, the
network does not need to squeeze all the useful past information into
the state variable, the cell state helps to save information from
the distant past. The cooperation of the internal memory and the external
memory outperforms vanilla RNN in a lot of real-world problem tasks
such as language translation \cite{sutskever2014sequence} video prediction
\cite{lotter2016deep,srivastava2015unsupervised} and so on. However,
all these sequence learning models have difficulties to accomplish
some simple memorization tasks such as copying and reversing a sequence.
The problem for LSTM and its variants is that the previous memories
are erased after they are updated, which also happens continuously
with the RNN state. In order to solve this problem, the concept of
online extracting events in time is necessary and can be conceptually
captured with an external memory bank. However, a learning system
with external event memory must also learn when to store an event,
as well as to use it in the future. In this way, the old memory does
not need to be erased to make space for the new memory. Neural stack
is an example which uses a stack as its external memory bank and gets
access to the stack top content through push and pop operations. Its
two operations are controlled by either a feedforward network or an
RNN such as \cite{jordan:attractor}\cite{hochreiter1997long}. The
research on neural stack network was originated in \cite{sun1993learning,sun2017neural}
to mimic the working mechanism of pushdown automata. The continuous
push and pop operations in it give an instruction of how to render
discrete data structure continuous to all the subsequent papers. The
authors in \cite{joulin2015inferring,grefenstette2015learning} adapted
these operations to queue and double queue to accomplish more complex
tasks, such as sequence prediction. Moreover, \cite{joulin2015inferring}
extended the number of stacks in their model. Recently, more powerful
network structures such as neural Turing machine\cite{graves2014neural}
and differentiable neural computer \cite{graves2016hybrid} are proposed.
In these network, all the contents in the memory bank can be accessed.
At the same time, Weston \cite{DBLP:journals/corr/WestonCB14} also
proposed a sequence prediction method using an addressable memory
and test it on language and reasoning tasks. Since the accessible
content is not restricted to the top of the memory as neural stack,
neural RAM has more flexibility to handle its memory bank. 

Many memory networks have emerged recently, some of them adopt internal
memory; some of them adopt external memory; some of them adopt logic
gates; some of them adopts a attention mechanism. As expected, all
of them have advantages for some specific tasks, but it's hard to
decide which one is optimal for a new task unless we have a clear
understanding of functions of all the components in the memory networks.
Although we can try and test them one by one from simple to complex,
it is a really time consuming process. And it is also not a good choice
to always go for the most complex network because it needs more resources
and takes more time to be well-trained. Intuitively, we all know that
if the network involves more components, it can make use of more information,
but what kinds of the extra information they are using and how useful
this extra information is, are still unknown in the current literatures.
Understanding the essential differences and relations between these
memory networks and connecting these differences to the requirements
of different targets is the key to choose the right network. Furthermore,
it can instruct us to design an appropriate memory network according
to the features of the specific sequences to be learned. 

In this paper, we analyze the capabilities of different memory networks
based on how they organize their memories. Moreover, we propose a
memory network taxonomy which covers the four main classes of memory
networks: vanilla RNN, LSTM, neural stack, neural RAM respectively
in our paper. Here, each class of network has several different realizations,
since the basic idea of all the variants are the same, we only choose
one typical network architecture for each class to do analyzation.
Our conclusion is that there is an hierarchical organization in the
sense that each one of them can be seen as a special case of another
one given the following order, i.e. vanilla RNN$\subseteq$ LSTM$\subseteq$neural
stack$\subseteq$neural RAM. This is no surprise, because it resembles
the hierarchical organization of language grammars, but here we are
specifically interested in linking the mapping architecture of the
learning machine to its descriptive power, which was not addressed
before. This inclusion relation is both proved mathematically and
verified with four synthetic tasks: counting, counting with interference,
reversing and repeat copying whose requirements of the past information
are increasing.

Our paper is organized as follows: after this introduction Section\ref{sec:intro},
Section \ref{sec:model} outlines the architecture of the four classes
of memory networks. Section \ref{sec:taxonomy} describes the proposed
taxonomy for memory networks and discussed how they organize their
memories, it also describes the four tasks developed to test the capabilities
of different networks. In Section \ref{sec:simu}, the proposed taxonomy
is corroborated by conducting four test experiments. Section \ref{sec:simu}
gives the conclusion and discuss some future works.

\section{Model}

\label{sec:model}

In this section, we introduce four typical network architectures which
represent four classes of memory networks adopted in this paper: vanilla
RNN, LSTM, neural stack and neural RAM. Their basic architectures
and training methods would be described in detail. 

\subsection{Vanilla RNN}

\begin{figure}
\includegraphics[scale=0.4]{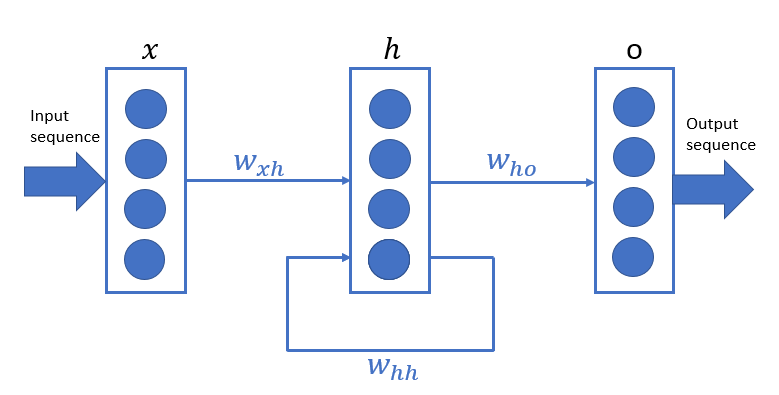}

\caption{Vanilla RNN}

\label{rnn}
\end{figure}

The RNN network \cite{jordan:attractor} is composed of three layers:
input, hidden recurrent and output layer. Besides all the feed forward
connections, there is a feedback connection from the hidden layer
to itself. The number of neurons in each layer is $K_{i},\,K_{h},\,K_{o}$.
The architecture of it is shown in Fig.\ref{rnn}. 

Given a sequence of symbols, $S:\:s_{1},\,s_{2},...,s_{T}$, each
symbol is encoded as single vector and fed as input to the network
one at a time. The dynamic of the hidden layer can be written as,

\begin{equation}
\mathbf{h}_{t}=f(\mathbf{w}_{xh}^{\mathrm{T}}\mathbf{x}_{t}+\mathbf{w}_{hh}^{\mathrm{T}}\mathbf{h}_{t-1}+\mathbf{b}_{h}),\label{eq:rnnstate}
\end{equation}
where $\mathbf{x}_{t}$ is the input at time $t$, which is the encoding
vector of symbol $s_{t}$. $\mathbf{w}_{xh}$ is $K_{h}\times K_{i}$
weighting matrix from input layer to hidden layer and $\mathbf{w}_{hh}$
is $K_{h}\times K_{h}$ recurrent weight, and $\mathbf{b}_{h}$ is
the $K_{h}\times1$ bias. $f(x)$ is the nonlinear activation function
such as the sigmoid activation function $\frac{1}{1+e^{-x}}$. The
output at time $t$ is,

\begin{equation}
\mathbf{o}_{t}=f(\mathbf{w}_{ho}^{\mathrm{T}}\mathbf{h}_{t}+\mathbf{b}_{o}),\label{eq:output}
\end{equation}
where $\mathbf{o}_{t}$ is the $K_{o}\times1$ output vector, $\mathbf{w}_{ho}$
is the $K_{o}\times K_{h}$ output weights, $\mathbf{b}_{o}$ is the
$K_{o}\times1$ bias. RNN can be cascaded and trained with gradient
decent methods called real time recurrent learning (RTRL) and backpropagation
through time (BPTT) \cite{werbos1990backpropagation} . 

The memory of the past at time $t$ is encoded in the hidden layer
variable $\mathbf{h}_{t}$. Although the upper limit of the differential
entropy of $\mathbf{h}_{t}$ is always larger than the total entropy
of the past input $\mathbf{x}_{1},\mathbf{x}_{2},...,\mathbf{x}_{t-1}$,
(i.e. the information is always smaller) the information captured
is highly dependent on the network weight. Even apart of the vanishing
gradient problem\cite{pascanu2013difficulty}, there is always a compromise
between memory depth and memory resolution in the RNN. In particular
for very long memory depths, the information is spread amongst many
samples and so there is a chance of overlaps amongst many past events. 

\subsection{LSTM}

\begin{figure}
\includegraphics[scale=0.35]{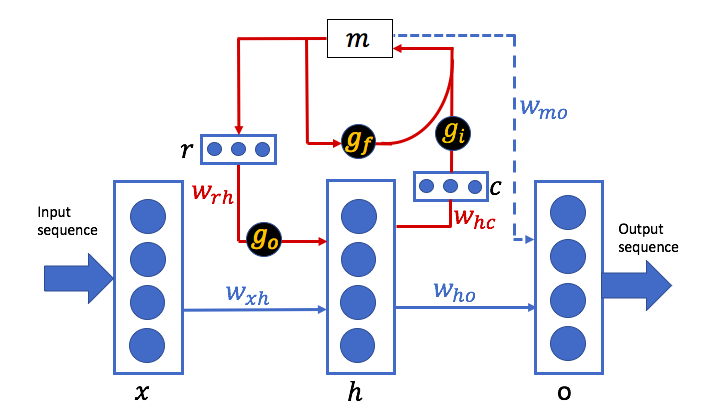}

\caption{LSTM}

\label{LSTM}
\end{figure}

\begin{figure}
\includegraphics[scale=0.3]{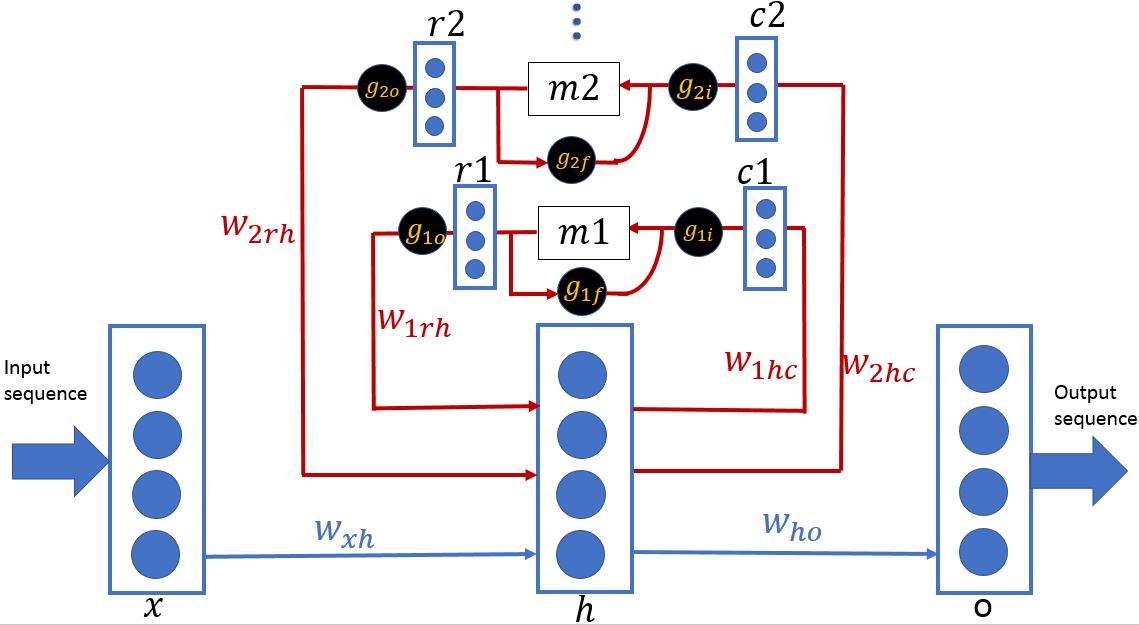}

\caption{LSTM with parallel memory slots}

\label{parallel LSTM}
\end{figure}

In LSTM\cite{hochreiter1997long}, as shown in Fig.\ref{LSTM}, the
feedback connection is a weighted vector $\mathbf{m}$ of the current
state and a long term state. The feedback method is described in Eq.(\ref{eq:lstm1})
to Eq.(\ref{eq:lstm4}). 

\begin{equation}
\mathbf{c}_{t}=f(\mathbf{w}_{hc}^{\mathrm{T}}\mathbf{h}_{t-1}+\mathbf{b}_{c}),\label{eq:lstm1}
\end{equation}

\begin{equation}
\mathbf{m}_{t}=g_{i,t}\mathbf{c}_{t}+g_{f,t}\mathbf{m}_{t-1},\label{eq:lstm2}
\end{equation}

\begin{equation}
\mathbf{r}_{t}=\mathbf{m}_{t},\label{eq:lstm3}
\end{equation}

\begin{equation}
\mathbf{h}_{t}=f(\mathbf{w}_{xh}^{\mathrm{T}}\mathbf{x}_{t}+\mathbf{w}_{rh}^{\mathrm{T}}g_{o,t}\mathbf{r}_{t}+\mathbf{b}_{h}),\label{eq:lstm4}
\end{equation}

where $g_{i,t}$, $g_{f,t}$, $g_{o,t}$ is the input gate, forget
gate and output gate at time $t$ respectively. $\mathbf{m}_{t}$
is the $N\times1$ long term memory at time $t$ which is initialized
as zero, $\mathbf{c}_{t}$ is the $N\times1$ candidate vector to
put into the long term memory and $\mathbf{w}_{hc}^{\mathrm{}}$ is
the corresponding weight of size $N\times K_{h}$. Different from
the vanilla RNN, the memory of the past is a combination of the long
term memory $\mathbf{m}_{t-1}$ and current state variable $\mathbf{c}_{t}$.
The weights between these two kinds of memories are decided by two
gates: forget gate $g_{f,t}$ decides how relevant the long term memory
is and the input gate $g_{i,t}$ decides how relevant the current
state is. Moreover, whether the calculated memory affects next state
is decided by the output gate $g_{ot}$, these three gates are calculated
as follows:

\begin{equation}
g_{i,t}=s(\mathbf{w}_{hg_{i}}^{\mathrm{T}}\mathbf{h}_{t-1}+\mathbf{w}_{xg_{i}}^{\mathrm{T}}\mathbf{x}_{t}+\mathbf{b}_{g_{i}}),\label{eq:gate1}
\end{equation}

\begin{equation}
g_{f,t}=s(\mathbf{w}_{hg_{f}}^{\mathrm{T}}\mathbf{h}_{t-1}+\mathbf{w}_{xg_{f}}^{\mathrm{T}}\mathbf{x}_{t}+\mathbf{b}_{g_{f}}),\label{eq:gate2}
\end{equation}

\begin{equation}
g_{o,t}=s(\mathbf{w}_{hg_{o}}^{\mathrm{T}}\mathbf{h}_{t-1}+\mathbf{w}_{xg_{o}}^{\mathrm{T}}\mathbf{x}_{t}+\mathbf{b}_{g_{o}}).\label{eq:gate3}
\end{equation}

where $\mathbf{w}_{hg_{i}}$, $\mathbf{w}_{hg_{f}}$, $\mathbf{w}_{hg_{o}}$
are $K_{h}\times1$ weights , $\mathbf{w}_{hf}$, $\mathbf{w}_{hf}$,
$\mathbf{w}_{hf}$ are $K_{i}\times1$ weights and $b_{g_{i}}$, $b_{g_{i}}$
and $b_{g_{i}}$ are bias. These three gates give flexibility to \textcolor{black}{operate
on memories}. For example, the memory in the long past can be obtained
by setting the forget gate as 1 and input gate as 0 for several consecutive
time steps. However, when the memory $\mathbf{m}_{t}$ is updated
as in Eq.(\ref{eq:lstm2}), the old value $\mathbf{m}_{t-1}$ is erased.
Hence, for the tasks which need more than one previous memories, we
must use several feedback loops in parallel as shown in \ref{parallel LSTM}.

The output of the network are either the same as the one in RNN as
shown in Eq.(\ref{eq:output}) or a function of the external memory,

\begin{equation}
\mathbf{o}_{t}=t(\mathbf{w}_{mo}^{\mathrm{T}}\mathbf{m}_{t}+\mathbf{b}_{o}),\label{eq:output-1}
\end{equation}

\subsection{Neural Stack}

\begin{figure}
\includegraphics[scale=0.45]{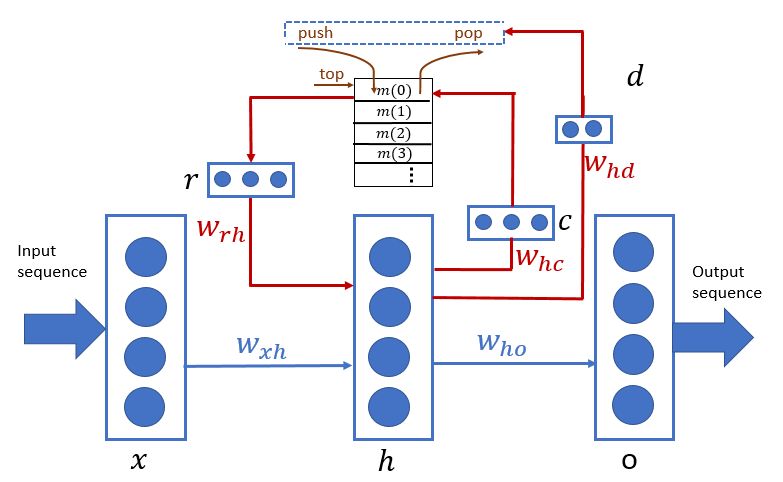}

\caption{Neural Stack}

\label{neuralstack}
\end{figure}

In this subsection, the neural network with an extra stack is introduced.
The diagram of the network is shown in Fig.\ref{neuralstack}. 

One stack property is that only the topmost content of the stack can
be read or and written. Writing to the stack is implemented by three
operations: push, adding an element to the top of the stack; pop,
removing the topmost of the stack; no-operation, keeping the stack
unchanged. These three operations can help the machine organize the
memory in a way to reduce the error. In order to train the network
with BPTT, all operations have to be implemented by continuous functions
over a continuous domain. According to \cite{sun2017neural,joulin2015inferring,grefenstette2015learning},
the domain of the operations are relaxed to any real value in $[0,\,1]$.
This extension adds an amplitude dimension to the operations. For
example, if the push signal $d_{push}=1$, the current vector will
be pushed into the stack as it is, if $d_{push}=0.8,$ the current
vector is first multiplied by 0.8 and then pushed onto the stack.
In this paper, the dynamics of the stack follows \cite{joulin2015inferring}.
To be specific, elements in stack would be updated as follows,

\begin{equation}
\mathbf{s}_{t}(0)=d_{t}^{push}\mathbf{c}+d_{t}^{pop}\mathbf{s}_{t-1}(1)+d_{t}^{no-op}\mathbf{s}_{t-1}(0),\label{eq:neural stacktop}
\end{equation}

\[
\mathbf{s}_{t}(i)=d_{t}^{push}\mathbf{s}_{t-1}(i-1)+d_{t}^{pop}\mathbf{s}_{t-1}(i+1)++d_{t}^{no-op}\mathbf{s}_{t-1}(i),
\]

$\mathbf{s}_{t}(i)$ is the content of the stack at time $t$ in position
$i$, $s_{t}(0)$ is the topmost content, $\mathbf{c}$ is the candidate
content to be pushed onto the stack, $d_{t}^{push}$ , $d_{t}^{pop}$
and $d_{t}^{no-op}$ are push, pop and no-operation signals. 

Neural network interacts with the stack memory by $d_{t}^{push}$,
$d_{t}^{pop}$, $d_{t}^{no-op}$, $\mathbf{c}_{t}$ and $\mathbf{r}_{t}$.
$\mathbf{r}_{t}$ is the read vector at time $t$, $\mathbf{r}_{t}=g_{o}\mathbf{s}_{t}(0)$.
$d_{t}^{push}$, $d_{t}^{pop}$, $d_{t}^{no-op}$ and $\mathbf{c}_{t}$
are decided by the hidden layer outputs and the corresponding weights,

\[
\mathbf{d}=[d_{t}^{push},\,d_{t}^{pop},\,d_{t}^{no-op}]^{\mathrm{T}}=s(\mathbf{w}_{hd}^{\mathrm{T}}\mathbf{h}_{t}+\mathbf{b}_{op}),
\]
where $\mathbf{w}_{hd}^{\mathrm{}}$ is the $3\times K_{h}$ weights
and $\mathbf{b}_{op}$ is the $3\times1$ bias.

\[
\mathbf{c}_{t}=g(\mathbf{w}_{hc}^{\mathrm{T}}\mathbf{h}_{t}+\mathbf{b}_{c}).
\]

Since the recurrence is introduced by the stack memory,

\begin{equation}
\mathbf{h}_{t}=g(\mathbf{w}_{xh}^{\mathrm{T}}\mathbf{x}_{t}+\mathbf{w}_{rh}^{\mathrm{T}}\mathbf{r}_{t}+\mathbf{b}_{h}),\label{eq:stackh}
\end{equation}

$\mathbf{r}_{t}$ is the read vector at time $t$, And the output
of the network is the same as (\ref{eq:output}). As all the variables
and weights are continuous, the error can be back propagated to update
the weights and bias.

With this external memory, all the useful information are retained.
Different from the internal memory, the content of past is not altered,
it is stored in its original form or the transformation form. What\textquoteright s
more, as the content and the operation of the past is separated, we
can efficiently select the useful content from this structured memory
other than using the mixture of all the content before. So the external
memory circumvents the compromise between the memory depth versus
memory resolution that is always present in the state memory. 

\subsection{Neural RAM}

\begin{figure}
\includegraphics[scale=0.4]{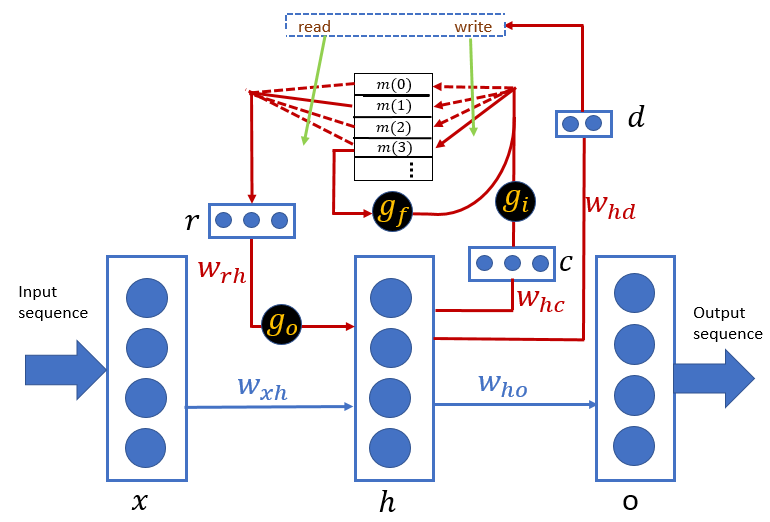}

\caption{Neural RAM}

\label{nerualram}
\end{figure}

The last and most powerful network is the Neural RAM as shown in Fig.\ref{nerualram}.
The neural RAM can be seen as an improvement of the neural stack in
the sense that all the contents in the memory bank can be read from
and written to. The challenge of the network is that all the memory
addresses are discrete in nature. In order to learn the read and write
addresses by error backpropogation \cite{rumelhart1986learning},
they have to be continuous. Papers \cite{DBLP:journals/corr/WestonCB14,graves2014neural,graves2016hybrid}
give a solution for this difficulty: reading and writing to all the
positions with different strengths. These strengths can also be explained
as the probabilities each position would be read from and written
to. One thing to note is that the read and write position do not need
to be the same ones. To be specific, the read vector at time step
$t$ is,

\begin{equation}
\mathbf{r}=\stackrel[i=0]{M-1}{\sum}a_{t}(i)\mathbf{m}_{t}(i),\label{neuralramra}
\end{equation}
$\mathbf{m}$ is the memory bank with $M$ memory locations, and $a_{t}(i)$
is the normalized weight for $i$th location at time $t$which satisfying,

\begin{equation}
\underset{i}{\sum}a_{t}(i)=1,\qquad0\leq a_{t}(i)\leq1.\label{eq:neuralrama}
\end{equation}
For the writing process, the forget and input gates arrangement is
also applicable here, for memory location $i$,

\begin{equation}
\mathbf{c}_{t}(i)=f(\mathbf{w}_{hc}^{\mathrm{T}}\mathbf{h}_{t-1}+\mathbf{b}_{c}),\label{eq:lstm1-1}
\end{equation}

\begin{equation}
\mathbf{m}_{t}(i)=g_{i,t}(i)\mathbf{c}_{t}(i)+g_{f,t}(i)\mathbf{m}_{t-1}(i),\!\forall i\label{eq:lstm2-1}
\end{equation}
here $g_{i,t}(i)$ and $g_{f,t}(i)$ together can be seen as the write
head for memory slot $i$ at time $t$. The dynamic of the hidden
layer is,

\begin{equation}
\mathbf{h}_{t}=f(\mathbf{w}_{xh}^{\mathrm{T}}\mathbf{x}_{t}+\mathbf{w}_{rh}^{\mathrm{T}}\mathbf{r}_{t}+\mathbf{b}_{h}),\label{eq:lstm4-1}
\end{equation}
Here the read weight $a_{t}(i)$ can be learned as, 
\begin{equation}
a_{t}(i)=f(\mathbf{w}_{ha}^{\mathrm{T}}\mathbf{h}_{t-1}+\mathbf{b}_{a}),\label{eq:readhear}
\end{equation}
 $\mathbf{w}_{ha}$ is the $M\times K_{h}$ weight. The nonlinear
activation function $f$ is usually set as softmax function. The write
weight and the output gate can be learned the same way as Eq.(\ref{eq:gate1})
and Eq.(\ref{eq:gate2}) in LSTM,

\begin{equation}
\mathbf{g}_{i,t}=s(\mathbf{w}_{hg_{i}}^{\mathrm{T}}\mathbf{h}_{t-1}+\mathbf{w}_{xg_{i}}^{\mathrm{T}}\mathbf{x}_{t}+\mathbf{b}_{g_{i}}),\label{eq:gate1-1}
\end{equation}

\begin{equation}
\mathbf{g}_{f,t}=s(\mathbf{w}_{hg_{f}}^{\mathrm{T}}\mathbf{h}_{t-1}+\mathbf{w}_{xg_{f}}^{\mathrm{T}}\mathbf{x}_{t}+\mathbf{b}_{g_{f}}),\label{eq:gate2-1}
\end{equation}

here $\mathbf{g}_{i,t}=[g_{i,t}(1),\,g_{i,t}(2),\,...,\,g_{i,t}(M)]^{\mathrm{T}}$,
$g_{f,t}=[g_{f,t}(i),\,g_{f,t}(2),\,...,\,g_{f,t}(M)]^{\mathrm{T}}$,
and $\mathbf{w}_{hg_{i}}$, $\mathbf{w}_{hg_{f}}$, $\mathbf{w}_{hg_{o}}$
are $K_{h}\times M$ weights , $\mathbf{w}_{hf}$, $\mathbf{w}_{hf}$,
$\mathbf{w}_{hf}$ are $K_{i}\times M$ weights and $\mathbf{b}_{g_{i}}$,
$\mathbf{b}_{g_{i}}$ and $\mathbf{b}_{g_{i}}$ are $M\times1$ bias.
In practice, instead of learning the read and write head from scratch,
some methods were proposed to simplify the learning process. For example
in Neural Turing machine \cite{graves2014neural}, $g_{ft}(i)$ is
coupled with $g_{it}(i)$ , $g_{ft}(i)=1-g_{it}(i)$. And read weight
$a_{t}$ and write weight $g_{it}$ are obtained by content-addressing
and location-addressing mechanisms. The content-addressing mechanism
gives the weights $a_{t}(i)$ (or $g_{it}(i)$) by checking the similarity
of the key $\mathbf{d}$ with all the contents in the memory, the
normalized version is,

\[
a_{t}(i)=\frac{exp(\alpha K[\mathbf{d},\mathbf{m}_{t}(i)])}{\sum_{j}(exp(\alpha K[\mathbf{d},\mathbf{m}_{t}(j)]))},
\]
where $\alpha$ is the parameter to control the precision of the focus,
$K$ is a similarity measure. Then, the weights will be further adapted
by the location-addressing mechanism. For example, the weights obtained
by content addressing can firstly blend with the previous weight and
then shfited for several steps,

\[
a_{t}(i)=g_{t}a_{t-1}(i)+(1-g_{t})a_{t}(i),
\]

\[
a_{t}(i)=a_{t}([i-n]_{M}).
\]
$g_{t}$ is the gate to balance the previous weight and current weight,
$n$ is the shifting steps, $[i-n]_{M}$ means the circular shift
for $M$ entities. Since the shifting operation is not differentiable,
the method in \cite{graves2014neural} should be utilized as an approximation. 

Another example is \cite{graves2016hybrid} which improves the performance
even more. To be specific, for reading, a matrix to remember the order
of memory locations they are written to can be introduced. With this
matrix, the read weight is a combination of the content-lookup and
the iterations through the memory location in the order they are written
to. And for writing, a usage vector is introduced, which guides the
network to write more likely to the unused memory. With this modification,
the neural RAM gets flexibility similar to working memory of human
cognition which makes it more suitable to intelligent prediction.
With these modifications, the training time for the neural RAM is
also reduced.

It should be pointed out that the RAM network can be seen as a LSTM
with several parallel feedback loops which coupled together in a non-trivial
way.

\section{A Unifying Memory Network Framework }

\label{sec:taxonomy}

\subsection{Memory network taxonomy}

\begin{figure}
\includegraphics[scale=0.5]{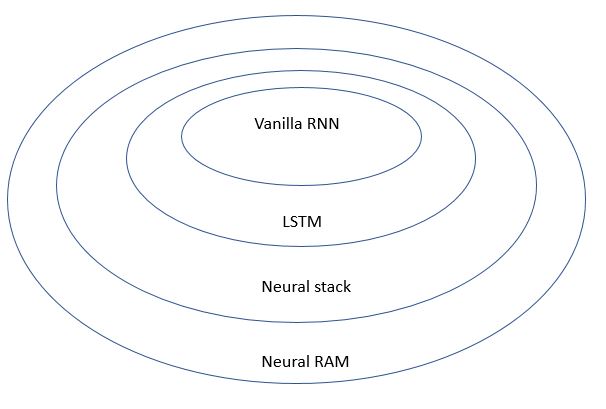}

\caption{Memory Network Taxonomy}

\label{hierarchy}
\end{figure}

In this section, we propose a hierarchy developed according to the
way the four kinds of memory described in the last section as shown
in Fig.\ref{hierarchy}. A network in the outer circle can always
implement a network in the inner circle as a special case; however,
the network in the inner circle will not have the capability to implement
the network in the outer circle's functionality, and so it will display
poorer performance. This hierarchy can help us choose the proper network
for a specific task. Our principle is always to choose the simplest
network since the complex network needs more resources. In this section,
we would first prove the inclusion relationship and then visualize
how these networks organize their memory space. 

\subsection{Inclusion Relationship Derivations}

In this subsection, we will prove neural stack is a special case of
neural RAM, LSTM is a special case of neural stack and vanilla RNN
is a special case of LSTM.

\subsubsection{From Neural RAM to Neural Stack}

Neural RAM is more powerful than neural stack because it has access
to all the contents in the memory bank. If we restrict the read and
write vector, neural RAM is degraded to neural stack. To be specific,
for the read head $a_{t}$, if all the read weights except the topmost
are set to zeros, then,

\begin{equation}
a(i)=\begin{cases}
0 & if\,i\neq0\\
t(\mathbf{w}_{ha}^{\mathrm{T}}\mathbf{h}_{t-1}+b_{a}), & if\,i=0
\end{cases},\label{eq:readhead1}
\end{equation}
here $\mathbf{w}_{ha}^{\mathrm{}}$ is $1\times K_{h}$ vector and
$b_{a}$ is the scalar.

And in the writing process, instead of learning all the contents to
be written to the stack as in Eq.(\ref{eq:lstm1-1}), only the content
to be put into M0 is learned as 

\begin{equation}
\mathbf{c}_{t}(0)=t(\mathbf{w}_{hc}^{\mathrm{T}}\mathbf{h}_{t-1}+\mathbf{b}_{c})+\alpha\mathbf{c}_{t-1}(1),\label{eq:ramtostack1}
\end{equation}
 all other contents are calculated as,
\begin{equation}
\mathbf{c}_{t}(i)=\alpha\mathbf{c}_{t-1}(i-1),\;if\;i\neq0.\label{eq:ramtostack2}
\end{equation}

Finally only the input and forget gates for the topmost element are
learned, all others just copy the values of the topmost's gates,
\begin{equation}
g_{i,t}(i)=g_{i,t}(0),\label{eq:gatei}
\end{equation}
\begin{equation}
g_{f,t}(i)=g_{f,t}(0).\label{eq:gatef}
\end{equation}

Since Eq.(\ref{eq:readhead1}) is a special case of Eq.(\ref{eq:readhear}),
Eq.(\ref{eq:ramtostack1}) , Eq.(\ref{eq:ramtostack2}) can be seen
as a special case of Eq.(\ref{eq:lstm1-1}), Eq.(\ref{eq:gatei})
and Eq.(\ref{eq:gatef}) can be seen as a special case of Eq.(\ref{eq:gate1-1})
and Eq.(\ref{eq:gate2-1}), hence the neural stack can be treated
as a special case of neural RAM. Here $a_{t}(0)$ works as the output
gate, $g_{i}(0),$ $\alpha g_{i}(0)$ and $g_{f}(0)$ works as the
push, pop and no-operate operations respectively. 

\subsubsection{From Neural Stack to LSTM }

According to Eq.(\ref{eq:lstm4}) and Eq.(\ref{eq:stackh}), the dynamics
of the neural stack have similar form as LSTM except for the reading
vector, i.e., the reading vector is $\mathbf{r}_{t}=g_{o}\mathbf{s}_{t}(0)$.
If we set the pop signal as zero, $d_{t}^{pop}=0$, and no operation
on the stack contents except for the topmost elements is avaiable,
then 

\[
\mathbf{s}_{t}(0)=d_{t}^{push}\mathbf{c}+d_{t}^{no-op}\mathbf{s}_{t-1}(0),
\]

\[
\mathbf{s}_{t}(i)=0,\;if\,i\neq0.
\]

Since the $d_{t}^{push}$, $d_{t}^{no-op}$ are calculated in the
same way the input gate $g_{i,t}$ and forget gate $g_{f,t}$ are
calculated in LSTM as shown in Eq.(\ref{eq:gate1}) to Eq.(\ref{eq:gate2}),
$d_{t}^{push}$can be seen as the input gate and $d_{t}^{no-op}$
can be seen as the forget gate. In this manner, this is exactly how
the LSTM organizes its memory. Hence, it is proved that LSTM can be
seen as the special case of neural stack.

\subsubsection{From LSTM to RNN}

Compared to RNN, LSTM introduces an external memory and the gate operation
mechanism. So if we set the output gate $g_{o}=0$, input gate $g_{i}=1$
and the forget gate $g_{f}=0$ instead of learning from the sequences,
the dynamics of LSTM is degraded to RNN as follows,

\begin{align}
\mathbf{h}_{t} & =t(\mathbf{w}_{xh}^{\mathrm{T}}\mathbf{x}_{t}+\mathbf{w}_{rh}^{\mathrm{T}}g_{o}\mathbf{r}_{t}+\mathbf{b}_{h})\label{eq:lstm}\\
 & =t[\mathbf{w}_{xh}^{\mathrm{T}}\mathbf{x}_{t}+\mathbf{w}_{rh}^{\mathrm{T}}\mathbf{r}_{r}+\mathbf{b}_{h}]\label{eq:a}\\
 & =t[\mathbf{w}_{xh}^{\mathrm{T}}\mathbf{x}_{t}+\mathbf{w}_{rh}^{\mathrm{T}}\mathbf{m}_{t}+\mathbf{b}_{h}]\nonumber \\
 & =t[\mathbf{w}_{xh}^{\mathrm{T}}\mathbf{x}_{t}+\mathbf{w}_{rh}^{\mathrm{T}}(g_{i}\mathbf{c}_{t}+g_{f}\mathbf{m}_{t-1})+\mathbf{b}_{h}]\nonumber \\
 & =t(\mathbf{w}_{xh}^{\mathrm{T}}\mathbf{x}_{t}+\mathbf{w}_{rh}^{\mathrm{T}}\mathbf{c}_{t}+\mathbf{b}_{h})\label{eq:b}\\
 & =t[\mathbf{w}_{xh}^{\mathrm{T}}\mathbf{x}_{t}+\mathbf{w}_{rh}^{\mathrm{T}}t_{1}(\mathbf{w}_{hc}^{\mathrm{T}}\mathbf{h}_{t-1}+\mathbf{b}_{c})+\mathbf{b}_{h}]\\
 & =t(\mathbf{w}_{xh}^{\mathrm{T}}\mathbf{x}_{t}+\mathbf{w}_{rh}^{\mathrm{T}}\mathbf{h}_{t-1}+\mathbf{b}_{h})\label{eq:c}
\end{align}

Here (\ref{eq:a}) is due to $g_{o}=0$, (\ref{eq:b}) is due to $g_{i}=1$
and $g_{f}=0$, (\ref{eq:c}) is because the weight $\mathbf{w}_{hc}$
and bias $\mathbf{b}_{c}$ are set as constants and the activation
function $t(x)$ is set as linear activation function,
\[
\mathbf{w}_{hc}=\mathbf{I},
\]
\[
\mathbf{b}_{c}=\mathbf{0},
\]
\[
t_{1}(x)=x.
\]

Since Eq.(\ref{eq:lstm}) is the dynamic of LSTM and Eq.(\ref{eq:c})
is the dynamic of RNN, the argument that RNN is a special case of
LSTM is proved.

\subsubsection*{Remark}

From the derivation we can draw a conclusion that, the innovation
of LSTM is the incorporation of an external memory and three gates
to balance the external memory and internal memory; the innovation
of Neural stack is to extend one external memory to several external
memories and to propose a method to visit the memory slots in a certain
order; the innovation of Neural RAM is to remove the constraint of
the memory visiting order, which mean any memory slot can be visited
at any time.

\subsection{Memory Space Visualization}

\begin{figure*}
\includegraphics{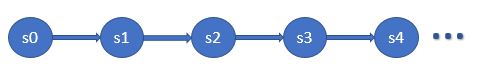}

\caption{vanilla RNN}

\label{rnnstate}
\end{figure*}

The analysis in this section ignores the influence of input. Fig.
\ref{rnnstate} shows the state transition diagram of RNN, where $s0,\,s1,...,s4$
represents the state at time $t_{0},\,t_{1},...,t_{4}$ respectively.
The blue arrow shows the variables' dependency relationship. For example,
state $s1$ is decided by $s0$, $s2$ is decided by $s1$ and so
on. The vanilla RNN has an \textcolor{black}{hidden Markov assumption}
for its input sequences, in other words, the current state can be
decided if the previous state is given. Thus for \textcolor{black}{Markov
sequence}, the RNN always performs well. However, for a lot of sequences
we need to deal with, the Markov assumption is not valid. In this
situation, the past memory helps a lot when we need to decide what's
the next state is. LSTM is the architecture which firstly take this
memory into consideration. 

\begin{figure*}
\includegraphics[scale=0.6]{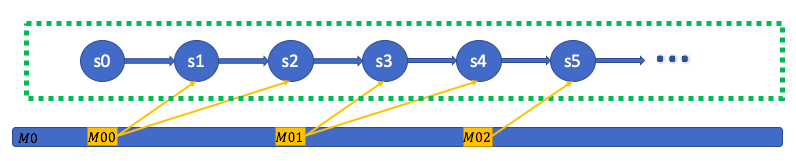}

\caption{LSTM}

\label{lstmstate}
\end{figure*}

In Fig.\ref{lstmstate}, a blue belt named $M0$ is used to save previous
memory: at $t_{0}$, \textcolor{black}{memory} $M00$ is generated
and saved, at time $t_{2}$, $M00$ is updated to $M01$ and at time
$t_{4}$, $M01$ is updated to $M02$. Every state is decided by its
previous one state and some older memory. The weight between these
two kinds of past information are decided by the input and forget
gates. For example, $s2$ is decided by $s1$ and $M00$, $s4$ is
decided by $s3$ and $M01$. A property of the memory is to forget
the older memory after it is updated. For instance, at time $t_{2}$,
when the memory is updated from $M00$ to $M01$ $M00$ is forgotten.
Thus, for the future state $s3,\,s4,\,s5,...$, they don't have access
to memory $M0$. According to this property, this architecture is
extremely useful when the previous states don't need to be addressed
again when they are updated. The capability of LSTM is greater than
vanilla RNN. Actually, RNN is LSTM without the memory belt. In other
words, LSTM is a RNN if the previous memory is not used (forget gate
is 0 and input gate is 1), as shown in the green dashed block in Fig.\ref{lstmstate}.

\begin{figure*}
\includegraphics{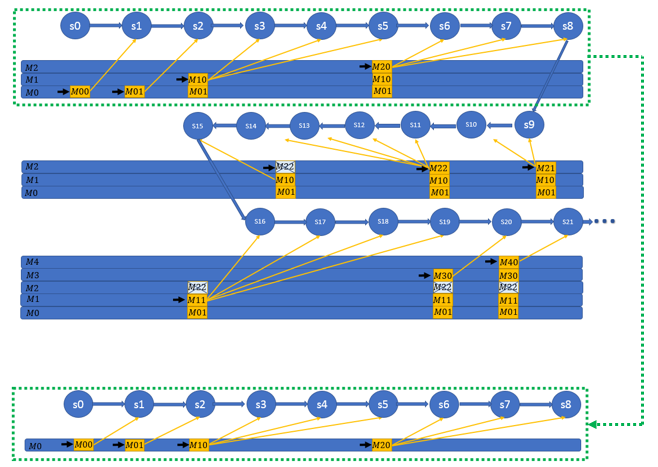}

\caption{Neural Stack}

\label{stackstate}
\end{figure*}

A more advanced memory neural network is the neural stack. It should
be mentioned that, in the current literature, there is no forget and
input gates embedded in the neural stack structure which makes it
work worse for some specific tasks. But these gates can be added into
the network the same way as LSTM. The push and pop operation provide
a way to save and address the previous memory as shown in Fig.\ref{stackstate}.
Let's assume, the network first saves state $M00$ in belt $M0$ and
updates it to $M01$. At time $t_{1}$, instead of replacing $M01$
with a new state $M10$, a new belt $M1$ is created to save $M10$.
In this way, both $M01$ and $M10$ are kept. Similarly, at time $t_{5}$,
$M20$ is saved in another belt $M2$. In time $t_{5}$, the content
in the stack is $M01$, $M10$, $M20$ and $M20$ is the topmost element.
Since all the useful past states are saved, they can be addressed
in the future. However, although it can go back to the previous memory,
it has two constraints. Firstly, it can not jump to any memory position,
the previous memory should be addressed and updated sequentially.
For example, as shown in the second line in Fig.\ref{stackstate},
if we want to go back to memory in belt $M1$, we have to go pass
memory in belt $M2$ first. Secondly, all the memory can only be accessed
twice, in other words, after the memory content is popped out of the
stack, it will be forgotten. For example, at time $t_{14}$, memory
in belt $M2$ is popped out, so in the future time step as shown in
the third line, content in belt $M2$ can not be accessed and updated
any more. 

LSTM can be seen as a special case of the neural stack if only the
push operation is allowed, as shown in the green dashed block in Fig.\ref{stackstate}.
Since all the contents in the stack below the topmost element will
never be addressed, only one belt is enough. Hence, the stack can
be squeezed to length 1 as shown Fig.\ref{stackstate}(b), which has
exactly the same structure as LSTM. 

From the state transition analysis above we can draw the conclusion
that, for the tasks where the previous memory need to be addressed
sequentially and at most twice, the stack neural network is our first
choice.

\begin{figure*}
\includegraphics[scale=0.9]{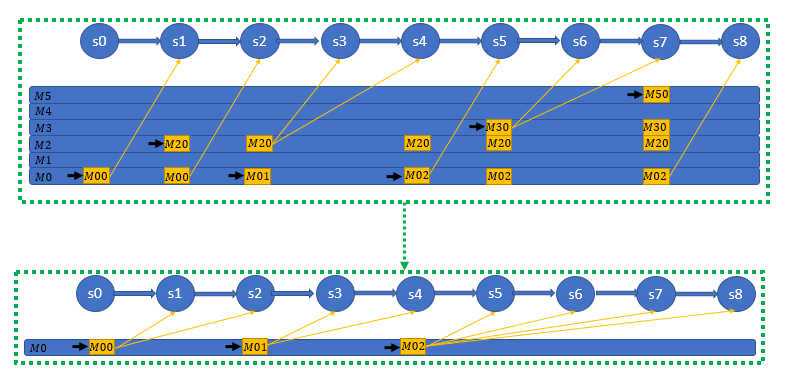}

\caption{Neural RAM}

\label{ramstate}
\end{figure*}

The most powerful memory access architecture in our hierarchy is the
neural RAM. Different from stack neural network, this kind of network
saved all the previous memory and can access any of them. There is
no requirement for the order of saving, updating and accessing the
memory. For example, in Fig.\ref{ramstate}(a), at time $t_{0}$,
memory $M00$ is saved in belt $M0$, at time $t_{1}$, it can directly
jump to belt $M2$. This neural RAM network can be degraded to the
stack if the order of memory saving and accessing is restricted as
shown in Fig. \ref{ramstate}(b). Hence, it can be further degraded
to the LSTM as shown in Fig.\ref{ramstate}(c).

All in all, since neural RAM can be degraded to neural stack, neural
stack can be degraded to LSTM, LSTM can be degraded to RNN, our inclusion
hierarchy is valid.

\subsection{Architecture Verification}

In order to verify the proposed taxonomy, 4 types of tasks using strings
of characters from easy to difficult are developed: counting, counting
with interference, reversing, repeat copying. 

For the counting task, the input sequence is the vector of $a$s and
the output sequence is the number of $a$s. For instance, when the
input sequence is $aaabcaa$, the output sequence would be 1233345.
For this kind of sequence, the state variable is needed to remember
the number of $a$s. As long as the network has the feedback loop,
the counting can be achieved. \textcolor{black}{Hence, vanilla RNN
is the best among all the memory network (In terms of error rate,
all of them have very small errors after training with enough time;
in terms of training speed, RNN outperforms all other method since
it uses least resources). This experiment also proves the argument
``LSTM is always better than RNN'' is not correct.}

For the counting with interference task, the input sequence are the
mixture of $a$, $b$ and $c$. We still want to count the number
of $a$, but if the input is $b$ or $c$, the output should also
be $b$ or $c$. For example, if the input is $aabbaca$, the output
sequence is $12bb3c4$. Assuming we have three neurons for both input
layer and output layer, the input sequence and output sequence after
a vector coding is,

\begin{tabular}{|c|c|c|}
\hline 
time step & input sequence & output sequence\tabularnewline
\hline 
\hline 
1 & {[}1 0 0{]} & {[}1 0 0{]}\tabularnewline
\hline 
2 & {[}1 0 0{]} & {[}2 0 0{]}\tabularnewline
\hline 
3 & {[}0 1 0{]} & {[}0 1 0{]}\tabularnewline
\hline 
4 & {[}0 1 0{]} & {[}0 1 0{]}\tabularnewline
\hline 
5 & {[}1 0 0{]} & {[}3 0 0{]}\tabularnewline
\hline 
6 & {[}0 0 1{]} & {[}0 0 1{]}\tabularnewline
\hline 
7 & {[}1 0 0{]} & {[}4 0 0{]}\tabularnewline
\hline 
\end{tabular}

For this kind of problem, an external memory cache is required, because
when $b$ or $c$ is encountered, the hidden layer's output value
will be overlaid. If we want to recall the memory of the number of
$a$s, it need to be saved in an external memory for future usage.
Thus the memory bank $\mathbf{m}$ in LSTM, neural stack and neural
RAM can work as this external memory. Hence, all of them except vanilla
RNN can complete the counting with interference task.

The third task is sequence reversing. For example. if the input sequence
is $abacde\delta------$, the output sequence should be $-------edcaba$.
\textcolor{black}{$\delta$ is the delimiter symbol, $-$ means any
symbol. When encountering $\delta$ in the input sequence, no matter
what the following symbols are, the output would be the input symbols
before $\delta$ in a reverse order. For this task, all the useful
past information should be stored and then retrieved in a reverse
order. Hence, the memory should have the ability to store more than
one contents and the read order is related to the write order. Since
RNN does not have this memory bank and LSTM's memory is forgotten
after it is updated, these two networks fails for this task. On the
other hand, both neural stack and neural RAM can save more than one
contents and the task satisfies the ``first in last out'' principle,
they can solve this task.}

The last task adopted here to verify the capability of networks is
the repeat copying task, by which we mean the output sequence is several
times a repetition of the input sequences. For example, if the input
sequence is $adbc\delta3--------------$ , the output should be $-------adbcadbcadbca\varepsilon$.
$\varepsilon$ is the end symbol. That is, when encountering the ending
signal $\delta$, the output will be the previous input sequence for
three times. For this kind of task, not only more than one past content
need be saved, they should be retrieved more than one time, here the
number is 3. In the neural stack, since all the saved information
is forgotten after being popped out, they can not be revisited again.
Thus, neural RAM is the only network that can handle this kind of
task.

These four synthetic sequences are good examples to show how the memory
networks operate on their respective memories to achieve a certain
goal. The details of the memory working mechanism is shown in the
simulation results in part \ref{sec:simu}.

\section{Experiment}

\label{sec:simu}

\subsection{Synthetic data}

In this section, preliminary experimental results would be presented
on four synthetic symbol sequences. The goal is to test the capability
of different memory networks and show how they organize their memories.
For all the experiments, four models are compared: vanilla RNN, LSTM,
neural stack, neural RAM.

\subsubsection{Counting }

\begin{figure}
\includegraphics[scale=0.45]{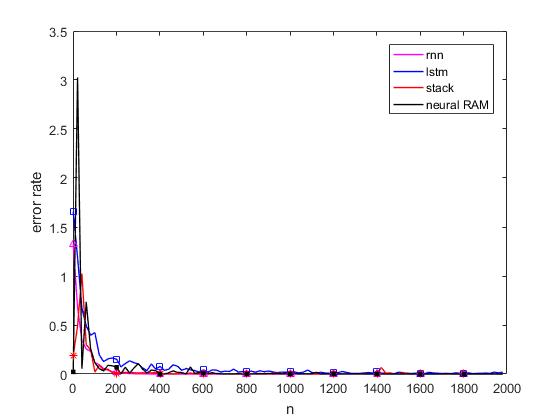}

\caption{Learning rate comparison for counting task}

\label{learningcurve}
\end{figure}

\begin{figure}
\includegraphics[bb=0bp 0bp 775bp 298bp,scale=0.28]{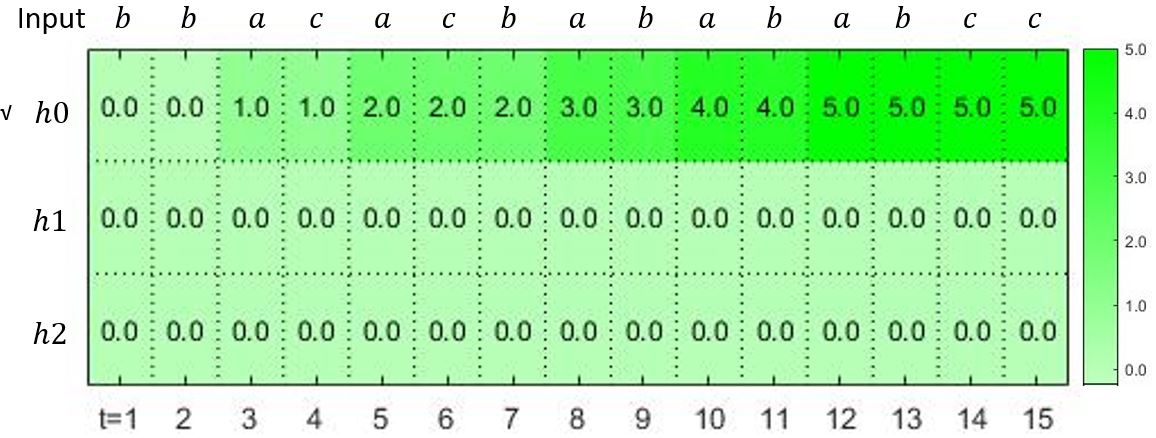}

\caption{Vanilla RNN: Internal memory content}

\label{MEMRNN}
\end{figure}

\begin{figure}
\includegraphics[scale=0.28]{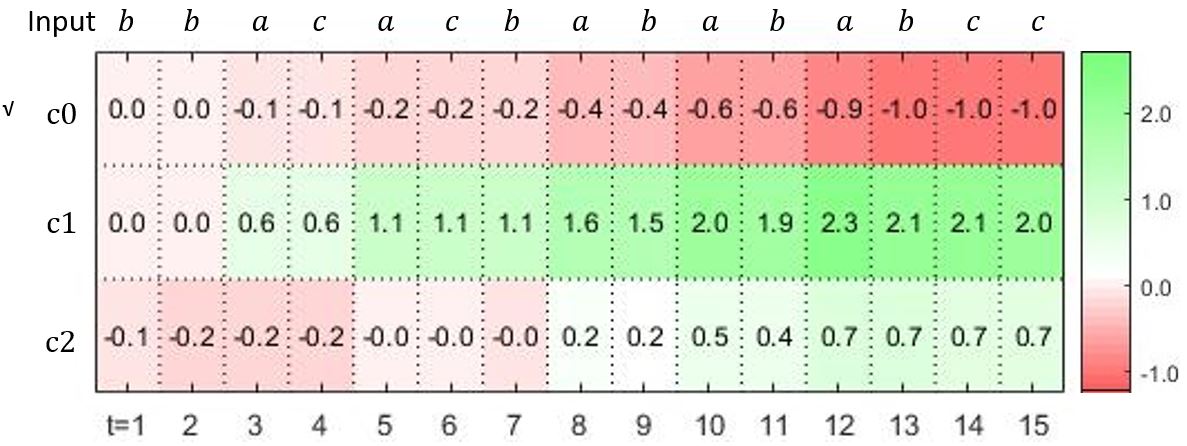}

\caption{LSTM: External memory content}

\label{MEMRNN-1}
\end{figure}

\begin{figure}
\includegraphics[scale=0.4]{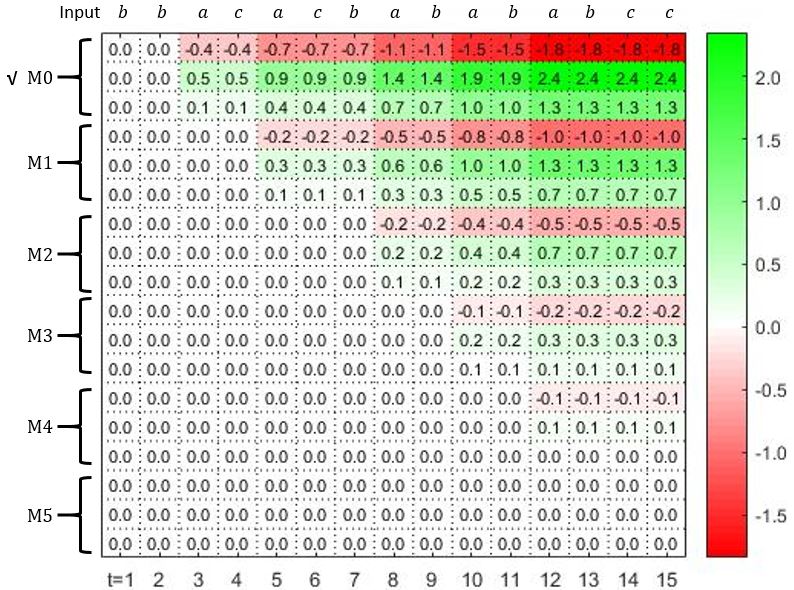}

\caption{Neural stack: stack content}

\label{MEMRNN-2}
\end{figure}

\begin{figure}
\includegraphics[scale=0.4]{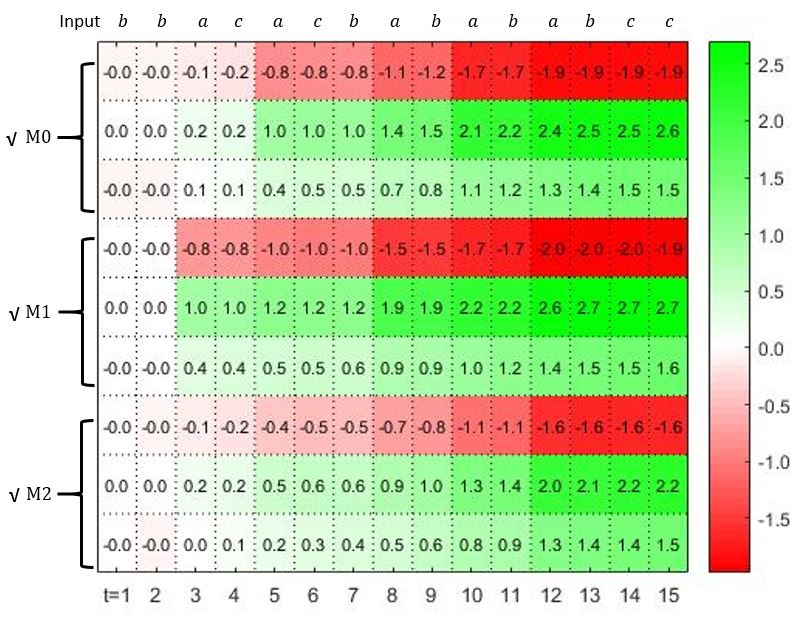}

(a) memory content

\includegraphics[scale=0.38]{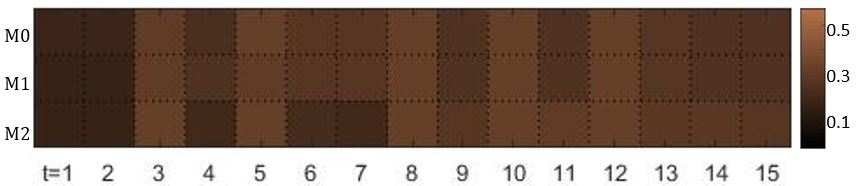}

(b) read

\includegraphics[scale=0.38]{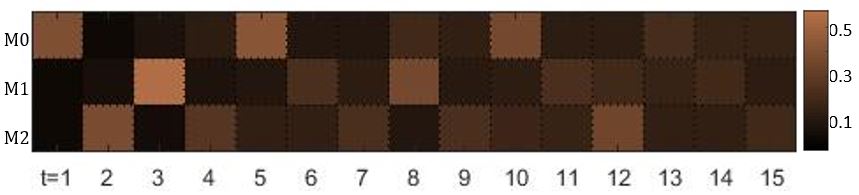}

(c)write

\caption{Neural RAM: Memory bank content and corresponding read and write operation}

\label{MEMRNN-3}
\end{figure}

From the analysis in section \ref{sec:taxonomy}, as long as the network
has a feedback loop to introduce memory of the past, it can count.
In this experiment sequences of symbols, $a,b,c$ are tested. All
the symbols are fed into the network one at a time as input vector
after a one-hot encoder and the network are trained with BPTT. 

In vanilla RNN, the activation function in the hidden layer is Relu
and the activation function in the output layer is sigmoid. In LSTM,
the external memory's content are initialized as zero. In the neural
stack, the push, pop and no-op operations are initialized as random
with mean 0 and variance 1. At first, there is only one content in
the stack which is initialized as zero. The depth of the stack can
increase to any number as required. In neural RAM, the word size and
memory depth are set as 3. The length of read and write vectors are
also set as 3. In LSTM, neural stack and neural RAM, the nonlinear
activation functions for all the gates are sigmoid functions and others
are tanh. The number of input neurons , hidden neurons and output
neurons are 3. All the weights are initialized as random variables
with mean 0 and variance 1, all the bias are initialized as 0.1. 

The model is trained with the synthetic sequences up to length 20.
When the input is $a$, the first elements in the output vector would
add one, otherwise, the output vector is unchanged.

The learning curve measured in mean square error (MSE) at the output
layer is shown in Fig.\ref{learningcurve}. From the results we can
see that, after 1000 training sequences, all the four models' errors
are less than 0.1. Fig.\ref{MEMRNN} to Fig\ref{MEMRNN-3} show the
details of the memory contents after the models are well-trained.
They are tested on a input sequence $bbacacbabababcc$ .Fig.\ref{MEMRNN}
shows that when $a$ is received, the first element of the hidden
layer is increased by 1. Fig.\ref{MEMRNN-2} shows that when receiving
$a$, the first element of the memory is decreased by 0.3, the second
and third elements have a similar pattern, but the increment is not
exactly a constant. However, as long as there is at least one element
in the memory learning the pattern, after multiplying with the weight
vector, the output of the network can give the expected values. Fig.\ref{MEMRNN-2}
shows how the neural stack uses its memory. Although the neural stack
has the potential to use unbounded number of stack contents, it only
uses the topmost content here, i.e. the push and no-operation cooperate
to learn the pattern. Fig.\ref{MEMRNN-2} shows the memory contents
of the neural RAM and the corresponding operations of it. From Fig.\ref{MEMRNN-3}(a),
we can see that all the three memory banks are learning the pattern,
hence, the read vector, all the three elements are all around 0.3
as shown in Fig.\ref{MEMRNN-3} (b). From this experiment we can see
that two memory banks are redundant here.

The goal of the counting experiment here is to show on the one hand
the capability of the four memory networks to remember one past state,
and on the other hand, to show the redundancy of the LSTM, neural
stack and neural RAM, i.e., the gate mechanism in LSTM, the unbound
number of the stack content in neural stack, and the multiple memory
contents in neural RAM. 

\subsubsection{Counting with interference}

\begin{figure}
\includegraphics[scale=0.45]{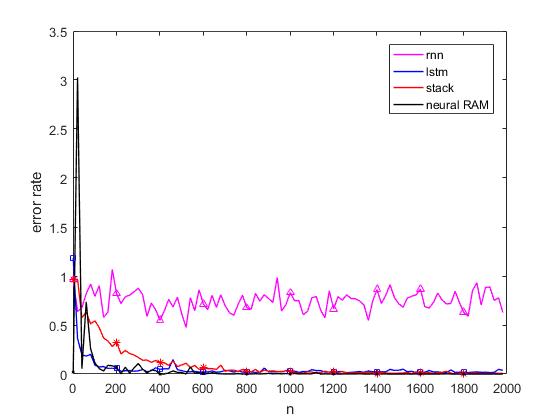}

\caption{Learning rate comparison for counting with interference task}

\label{learningcurve2}
\end{figure}

\begin{figure}
\includegraphics[scale=0.35]{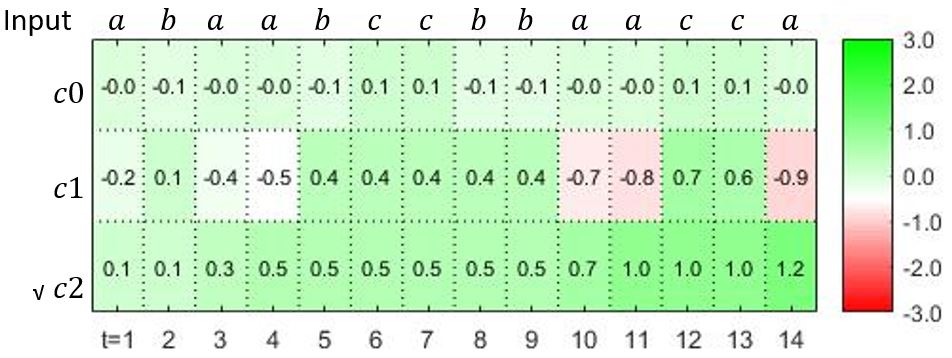}

\caption{LSTM: External memory content}

\label{task2mem}
\end{figure}

\begin{figure}
\includegraphics[scale=0.45]{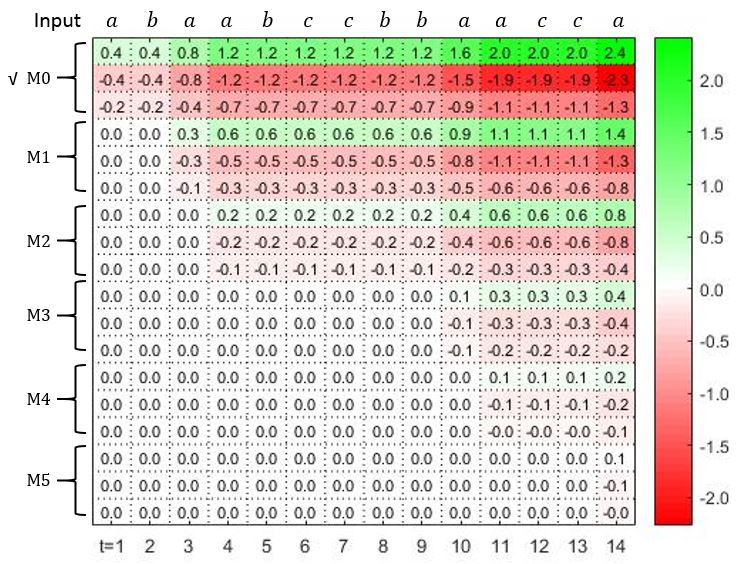}

\caption{Neural stack: stack content}

\label{task2mem-1}
\end{figure}
\begin{figure}
\includegraphics[scale=0.45]{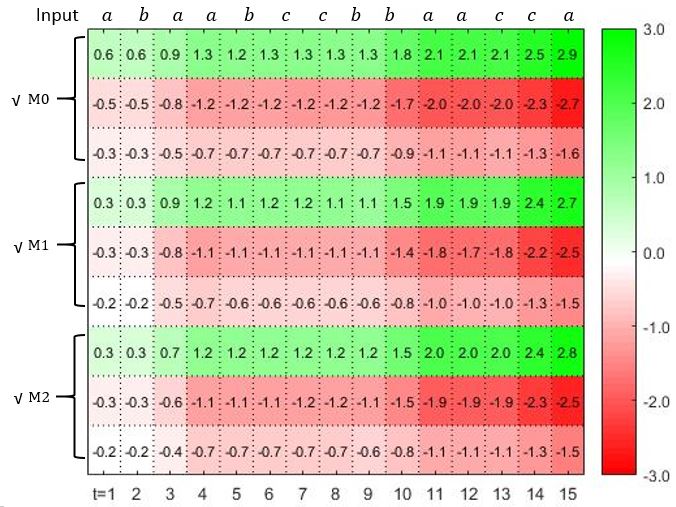}

(a) memory content

\includegraphics[scale=0.35]{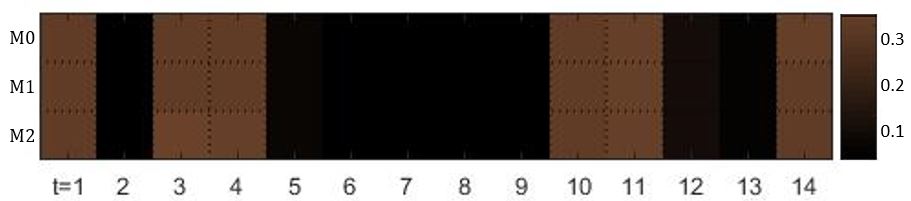}(b) read

\includegraphics[scale=0.35]{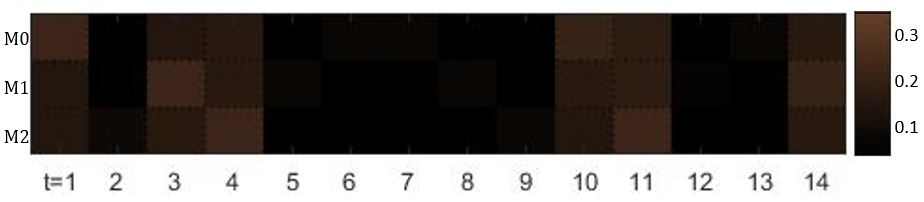}

(c)write 

\caption{Neural RAM: Memory bank content and corresponding read and write operation}

\label{task2mem-2}
\end{figure}

The second experiment is to test the external memory, i.e., the capability
of putting the memory aside and using it when it is needed. This is
the new feature of LSTM, neural stack and neural RAM implemented by
the gate mechanism. All the settings for the experiments are the same
as the counting task except when inputting $b$ and $c$. In the counting
task, the desired operation is the same as the one in the last time
step when inputting $b$ and $c$. Here, the desired response is $[0,\,1,\,0]$
when inputting $b$ and $[0,\,0,\,1]$ when inputting $c$. In order
to accomplish this task, the useful memory of the past should be put
aside and not disturbed. Since in vanilla RNN, the only memory of
the past is the internal memory, it will be refreshed when inputting
$b$ and $c$, vanilla RNN can never learn the pattern. Fig.\ref{learningcurve2}
shows the learning curve for these four networks, all the networks
except the vanilla RNN have errors less than 0.1 after enough training
samples, which is in consistent with our analysis.

Fig.\ref{task2mem} to Fig.\ref{task2mem-2} shows the memory usage
of LSTM, neural stack and neural RAM. Fig.\ref{task2mem} shows that
every time the symbol $a$ is input, the third element of the memory
content would increase by around 0.2. Fig.\ref{task2mem-1} and Fig.\ref{task2mem-2}
also show the similar incremental patterns of neural stack and neural
RAM. An notable difference between Fig.\ref{task2mem}-Fig.\ref{task2mem-1}
and Fig.\ref{MEMRNN-1}-Fig.\ref{MEMRNN-2} is the usage of the memory.
When dealing with counting task, the output gates are always 1, however,
when dealing with counting with interference task, the output gates
are 0 when inputting $b$ and $c$, this helps to cut off the interference
from the memory. Similarly, the read vector are always around {[}0.3,
0.3, 0.3{]} in Fig.\ref{MEMRNN-3}, however, in Fig. \ref{task2mem-2},
the read vector's elements are almost zeros when encountering $b$
and $c$. The read vector here works as the output gates in LSTM and
neural stack. It also shows why neural RAM does not need an output
gate. 

All in all, the goal of this experiment is to show the effect of the
gate mechanism and the redundancy of the unbound stack content in
neural stack and multiple memory banks in neural RAM.

\subsubsection{Reversing}

\begin{figure}
\includegraphics[scale=0.45]{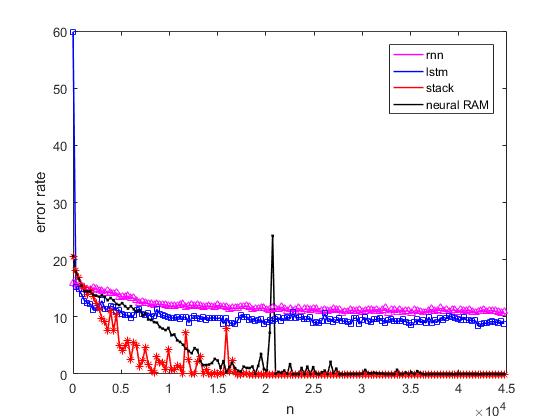}

\caption{Learning rate comparison for reversing task}

\label{learningcurve3}
\end{figure}

\begin{figure}
\includegraphics[scale=0.45]{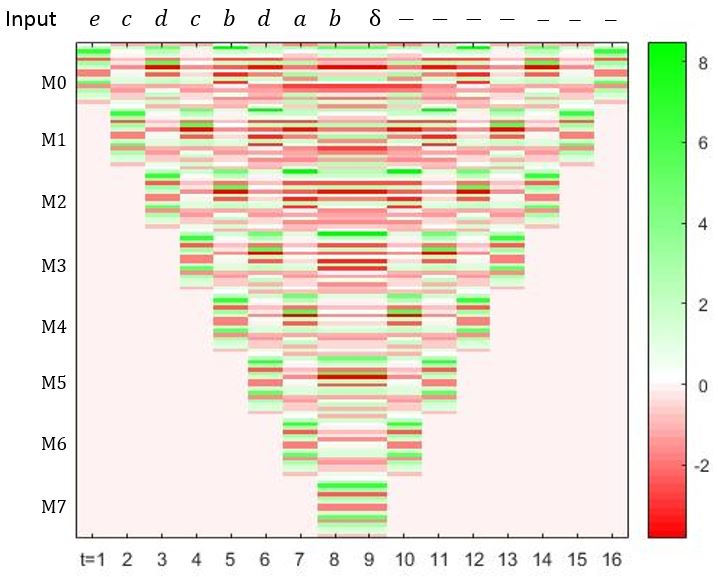}

\caption{Neural stack: stack content}

\label{task3mem}
\end{figure}

\begin{figure}
\includegraphics[scale=0.4]{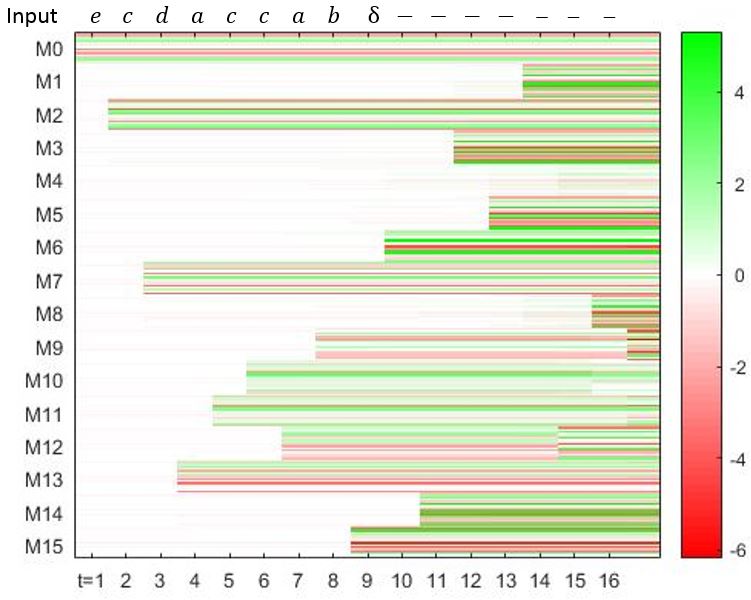}

(a) memory content

\includegraphics[scale=0.45]{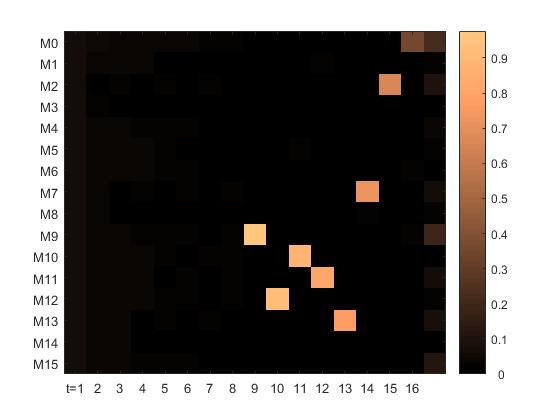}

(b) read

\includegraphics[scale=0.45]{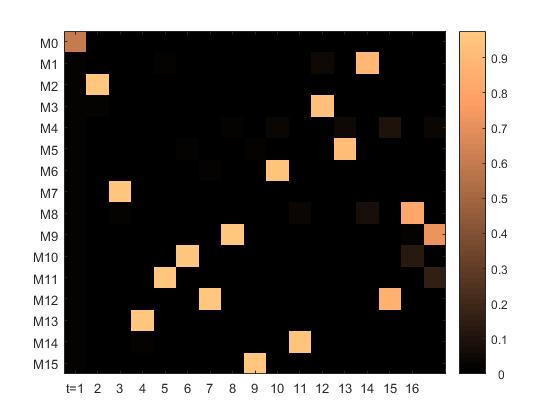}

(c) write

\caption{Neural RAM: Memory bank content and corresponding read and write operation}

\label{task3mem-1}
\end{figure}

For the sequence reversing problem, every symbol in the first half
of the sequence is randomly picked in the set \{$a,\,b,\,c,\,d,\,e$\},
then a \textcolor{black}{delimiter} symbol $\delta$ follows, and
the second half of the sequence is the reverse of the first half.
The performance is measured on the error rate in output entropy for
the second half.

In this experiment, some setting are different from the first two
experiments. In vanilla RNN, the activation function in the hidden
layer is sigmoid function since we use entropy instead of mean square
error as the cost function. In neural RAM, the word size and memory
depth are set as 16. The length of read and write vectors are also
set as 16. The number of input neurons, hidden neurons and output
neurons are 6, 64, 6. The model is trained with sequences up to length
20. To finish this task, all the input samples have to be saved in
the memory and be visited in the reverse order. Hence networks with
no external memory as RNN or a external memory as LSTM fail. Learning
curves of the four networks are shown in Fig.\ref{task3mem}. We can
see that vanilla RNN and LSTM do not have the capability of reversing
the sequence no matter how many samples are used for training. Fig.\ref{task3mem}
shows how neural stack utilizes its stack memory to solve this problem.
Since each memory bank's word size is 16, here we only use colors
instead of the specific numbers to show the values of contents in
memory. Different from the first two tasks, the function of the stack
is finally exploited. In the first half sequence, the input symbols
are encoded as 16-elements vectors and pushed into the stack. In the
second half of the sequence, the contents in the stack are popped
out sequentially. It should be noticed that as long as the the contents
are popped out, they can not be revisited anymore. Different from
neural stack, the contents in neural stack are never wiped as shown
in Fig.\ref{task3mem-1}. The contents in the memory banks are only
wiped if they are useless in the future or the memory banks are not
enough so they have to be wiped to make space for new stuffs. Another
feature of the memory bank for neural RAM is the memory banks are
not used in order such as M0, M1, M2...In this example, the memory
banks are used in the order M0, M2, M7, M13.... But as long as the
network knows the writing order, the task can be accomplished. Fig\ref{task3mem-1}(b)(c)
shows the reading and writing weights, we can see that the second
half of the reading weights is the mirror of the first half of the
sequence of the writing weights, which means the network learns to
reverse. 

Overall, the goal of this task is to show the advantage of the multiple
memory banks and how the neural RAM and neural stack organize their
memory banks.

\subsubsection{Repeat copying}

\begin{figure}
\includegraphics[scale=0.45]{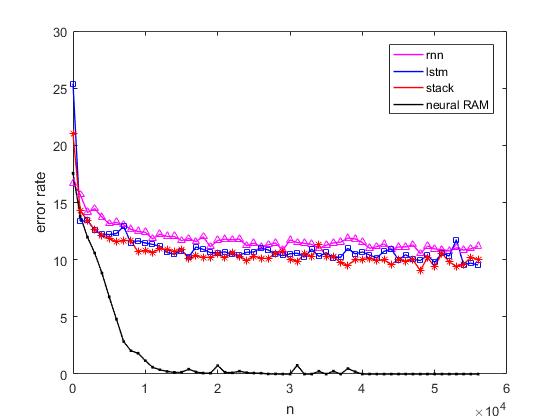}

\caption{Learning rate comparison for repeat copying task}

\label{learningcurve4}
\end{figure}

\begin{figure}
\includegraphics[scale=0.4]{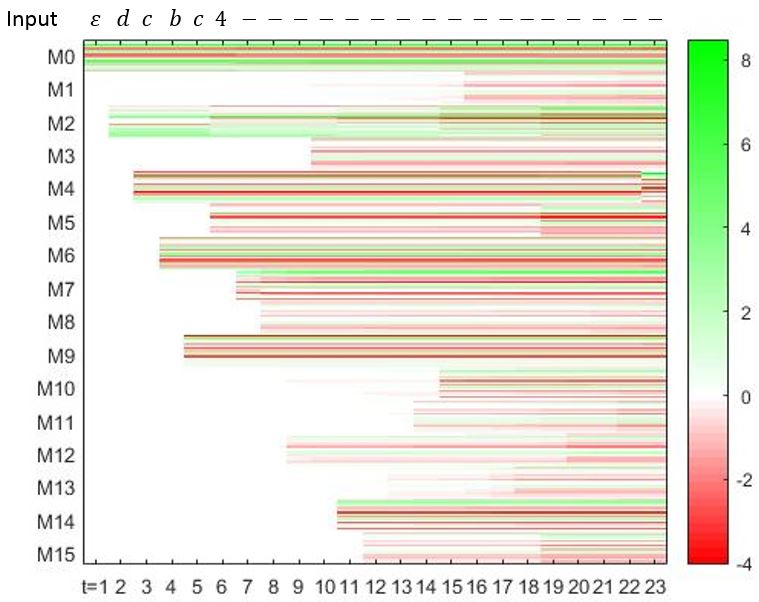}

(a) memory content

\includegraphics[scale=0.45]{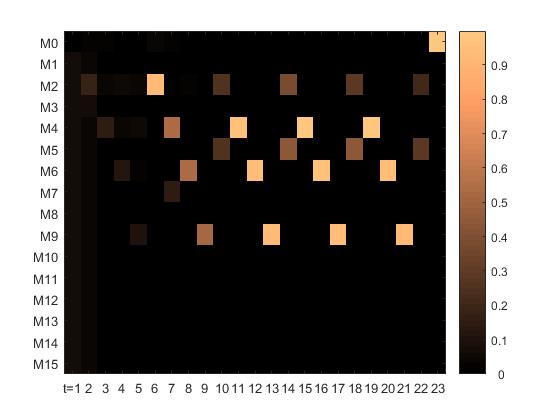}

(b) read

\includegraphics[scale=0.45]{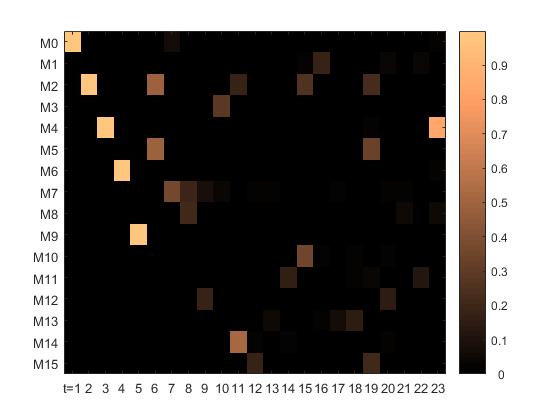}

(c) write

\caption{Neural RAM: Memory bank content and corresponding read and write operation}

\label{task4mem}
\end{figure}

The last and hardest problem we are going to implement is the repeat
copying task. To accomplish this task, the contents saved in the memory
banks can not be wiped when they are used for one time. Hence, neural
stack is the only network type which can handle this problem. In this
experiments, the training sequences are composed of a starting symbol
$\epsilon$, some symbols in set \{$a,\,b,\,c,\,d,\,e$\} followed
by a repeating number symbol $\delta$ and some random symbols. $\epsilon,\,a,\,b,\,c,\,d,\,e$
are one-hot encoded with on value 1 and off value 0; $\delta$ is
encoded with on value $n$ and off value 0, $n$ is the repeating
number. Fig.\ref{learningcurve4} shows the learning curves for the
four network and the fact that neural RAM is the only network that
can handle this problem. Fig.\ref{task4mem} shows how the neural
RAM solves this problem. From the writing weights we can see that,
the starting symbol is saved in M0, and symbols needed to be repeated
are save in M2, M4, M6, M9. After t=4, the network would read from
M2/M5, M4,M6,M9. At the beginning of every loop, the network reads
from both M2 and M5 probably because the repeating time symbol $\delta$
is saved in M5. The value in M5 can tell the network whether to continue
repeating or to output the ending symbol. We can see from Fig.\ref{task4mem}(b),
at time t=22, after reading from M2 and M5, the network stops reading
from M4 to M9 and turns to M0. 

The goal of this task is to test whether the networks can operate
their memory banks and show the advantage of neural RAM compared to
other memory networks.

\subsection{Real world problem}

In this section, we will use two natural language processing problems
to show the different capabilities of these four kinds of networks.
These two examples also give us some hints on how to choose the right
memory networks according to the specific tasks.

\subsubsection{Sentiment Analysis}

The first experiment is sentiment analysis problem, by which we mean
giving a paragraph of texts, determining whether the emotional tone
of the text is negative or positive. For example, an example from
lmdb movie review dataset with negative emotion is, 

``Outlandish premise that rates low on plausibility and unfortunately
also struggles feebly to raise laughs or interest. Only Hawn's well-known
charm allows it to skate by on very thin ice. Goldie's gotta be a
contender for an actress who's done so much in her career with very
little quality material at her disposal.''

And a positive text is,

``I absolutely loved this movie. I bought it as soon as I could find
a copy of it. This movie had so much emotion, and felt so real, I
could really sympathize with the characters. Every time I watch it,
the ending makes me cry. I can really identify with Busy Phillip's
character, and how I would feel if the same thing had happened to
me. I think that all high schools should show this movie, maybe it
will keep people from wanting to do the same thing. I recommend this
movie to everybody and anybody. Especially those who have been affected
by any school shooting. It truly is one of the greatest movies of
all time.''

The output of the neural network should be {[}1, 0{]} for the first
paragraph and {[}0, 1{]} for the second paragraph. After encoding
all the words into vectors, they are fed into the network one by one.
The decision of the tone of the paragraph will be made at the end
of the paragraph. Here we use a pretrained model: GloVe \cite{pennington2014glove}
to create our word vector. The matrix contains 400,000 word vectors,
each with a dimensionality of 50. The matrix is created in a way that
words having similar definitions or context reside in the relatively
same position in the vector space. The dataset adopted here is the
lmdb movie review data which has 12500 positive reviews and 12500
negative reviews. Here we use 11500 reviews for training and 1000
data for testing. In neural RAM, the word size and memory depth are
set as 64. The number of read and write head are 4 and 1. In LSTM,
neural stack and neural RAM, the nonlinear activation functions for
all the gates are sigmoid. The activation functions at the output
layer is sigmoid and others are tanh. The number of input neurons,
hidden neurons and output neurons are 50, 64, 2. 

In order to judge the emotional tone of the text as the end, an external
memory whose value would be affected by some key words is useful.
And since the goal here is to classify the emotional tone as either
1 or 0, the specific contents are not very important here so there
is no need to store all of them. Hence, the memory banks do not show
advantages here. Since LSTM has this external memory and it needs
less time to train compared to neural stack and neural RAM, it should
be best choice among these four kinds of networks for this task.

Table \ref{setiment} shows averaging error rates of 5 runs. We can
see that all the networks with external memories have similar performance,
which is in compliance with our analysis.

\begin{table}
\caption{Error rate for movie review}

\begin{tabular}{|c|c|c|c|c|}
\hline 
 & vanilla RNN & LSTM & neural Stack & neural RAM\tabularnewline
\hline 
\hline 
error rate & 31\textbf{$\pm$}5 & \textbf{19$\pm$2.5} & 23$\pm$10 & 20$\pm$9\tabularnewline
\hline 
\end{tabular}

\label{setiment}
\end{table}

\subsubsection{Question Answering}

In this section, we investigate the performance of these four networks
on three question answering tasks. The target is to give an answer
after reading a little story followed by a question. For example,
the story is 

``Mary got the milk there. John moved to the bedroom. Sandra went
back to the kitchen. Mary travelled to the hallway.''

And the question looks like, ``Where is the milk?''. The machine
is expected to give answer ``hallway''. For this problem, in order
to give the right answer, machine should memorize the facts that Mary
got the milk and travelled to the hallway. What's more, since the
machine doesn't know the question when reading the stories, it has
to store all the useful facts in the story. Thus a large memory bank
where all the contents can be visited is useful here. According to
our analysis in part \ref{sec:taxonomy}, neural RAM should perform
the best here.

In order to verify our conjecture, we test these four networks on
three tasks from bAbI dataset\cite{weston2015towards}. For each task,
we use the 10,000 questions to train and report the error rates on
the test set in Table \ref{qa}. In vanilla RNN and neural stack,
the nonlinear activation functions for all the gates are sigmoid.
The activation functions at the output layer is sigmoid and others
are tanh. The number of input neurons, hidden neurons and output neurons
are 150, 64, 150. The experimental settings for LSTM and neural RAM
are the same as \cite{graves2016hybrid} and the results for these
two networks are from \cite{graves2016hybrid}. From the results,
we can see that neural RAM achieves the best performance. On thing
to be mentioned here is, although the mean error rate of the neural
RAM is the lowest, the variance is larger than all others. We believe
the reason for this is the complexity of the network, which leads
to too many local minimal points. Since the point here is to check
the capabilities of different neural memory networks, a better way
to utilize the external memory and train the network will be our future
work.

\begin{table}
\caption{Error rate for three tasks from bAbI tasks }

\begin{tabular}{|c|c|c|c|c|}
\hline 
Task & vanilla RNN & LSTM & neural Stack & neural RAM\tabularnewline
\hline 
\hline 
1 supporting fact & 52$\pm$1.5 & 28.4$\pm$1.5 & 41$\pm$2.0 & \textbf{9.0$\pm$12.6}\tabularnewline
\hline 
2 supporting facts & 79$\pm$2.5 & 56.0$\pm$1.5 & 75$\pm$6 & \textbf{39.2$\pm$20.5}\tabularnewline
\hline 
3 supporting facts & 85$\pm$2.5 & 51.3$\pm$1.4 & 78$\pm$6.4 & \textbf{39.6$\pm$16.4}\tabularnewline
\hline 
\end{tabular}

\label{qa}
\end{table}

From these two examples, we can see that the taxonomy proposed in
this paper can helps us to analyze the properties of the specific
problem and choose the right neural network. But people still have
to analyze the problem by themselves. A data-driven method to analyze
the tasks and choose the appropriate network is our next step.

\section{Conclusion and future work}

\label{sec:con}

In this paper, we propose a taxonomy based on the state structure
for the memory networks recently proposed. The taxonomy are proved
mathematically and verified with simple synthetic sequences. Moreover,
this work only analyzes what tasks these networks can or can not do,
the next step is to analyze the performance of these network and explore
the method to improve the memory utilization efficiency. How to use
this taxonomy to design an appropriate network for some real-wold
problem is our future work.

\bibliographystyle{unsrt}
\nocite{*}
\bibliography{IEEEabrv,reference}

\end{document}